\documentclass{article}
\pdfoutput=1

\PassOptionsToPackage{numbers,compress,sort}{natbib}
\usepackage
  [final]
  {neurips_2023}

\usepackage[utf8]{inputenc} 
\usepackage[T1]{fontenc}    

\usepackage{gillius}
\usepackage[scaled=0.85]{FiraMono}

\usepackage{soul}
\usepackage[english]{babel}
\usepackage[babel,final,expansion=alltext,protrusion=alltext-nott]{microtype}

\usepackage{enumitem}
\usepackage{setspace}
\usepackage{multirow}
\usepackage{makecell}
\usepackage{graphicx}
\usepackage[hyphens]{url}
\usepackage{hyperref}
\usepackage{booktabs}
\usepackage{amsfonts}
\usepackage{nicefrac}
\usepackage[dvipsnames]{xcolor}
\usepackage[capitalize,nameinlink,noabbrev]{cleveref}

\newlist{andlist}{enumerate*}{1}
\setlist[andlist]{
  label=\bfseries(\alph*),
  itemjoin={{, }},
  itemjoin*={{, and }}
}

\setitemize{
  itemsep=0.5pt,
  leftmargin=2.5em,
  topsep=0.75pt
}

\setenumerate{
  itemsep=0.5pt,
  leftmargin=2.5em,
  topsep=0.75pt
}

\hypersetup{
  colorlinks = true,
  citecolor = Maroon,
  linkcolor = Plum,
  urlcolor = Blue,
}

\crefname{section}{\S}{\S}

\newcommand{\nChromebooks}{54}
\newcommand{\nDataPoints}{100K}
\newcommand{\nManufacturers}{4}
\newcommand{\nSOCVendors}{4}
\newcommand{\meanRScore}{$97.8\%$}
\newcommand{\meanMAAPE}{$10.1\%$}

\newcommand{\ccol}[1]{\multicolumn{1}{c}{\textls[50]{\textsf{#1}}}}

\title{Predicting User Experience on Laptops \\ from Hardware Specifications}

\author{%
  Saswat Padhi \\
  Google Inc.\\
  \texttt{spadhi@google.com} \\
  \And
  Sunil K. Bhasin \\
  Google Inc.\\
  \texttt{skbhasin@google.com} \\
  \And
  Udaya K. Ammu \\
  Google Inc. \\
  \texttt{udaykiran@google.com} \\
  \AND
  Alex Bergman \\
  Google Inc. \\
  \texttt{abergman@google.com} \\
  \And
  Allan Knies \\
  Google Inc.\\
  \texttt{aknies@google.com} \\
}

\begin{document}

\maketitle

\newcommand{\sfemph}[1]{\textsf{\textit{#1}}}

\begin{abstract}
  Estimating the overall user experience (UX) on a device is a 
  common challenge faced by manufacturers.
  Today, device makers primarily rely on microbenchmark scores,
  such as Geekbench, that stress test specific hardware components,
  such as CPU or RAM, but do not satisfactorily capture consumer workloads.
  System designers often rely on domain-specific heuristics
  and extensive testing of prototypes to reach a desired UX goal,
  and yet there is often a mismatch between the manufacturers' performance claims
  and the consumers' experience.

  We present our initial results on predicting real-life user experience
  on laptops from their hardware specifications.
  We target web applications that run on Chromebooks (ChromeOS laptops)
  for a simple and fair aggregation of experience across applications and workloads.
  On \nChromebooks\ laptops, we track 9 UX metrics on common end-user workloads:
  web browsing, video playback and audio\,/\,video calls.
  We focus on a subset of high-level metrics exposed by the Chrome browser,
  that are part of the Web Vitals initiative
  for measuring user experience on web applications.
  
  With a dataset of \nDataPoints\ UX data points,
  we train gradient boosted regression trees
  that predict the metric values from device specifications.
  Across our 9 metrics,
  we note a mean $R^2$ score (goodness-of-fit on our dataset) of \meanRScore\
  and a mean MAAPE (percentage error in prediction on unseen data) of \meanMAAPE.
\end{abstract}

\section{Introduction}%
\label{sec:introduction}

Computer hardware continues to evolve rapidly.
Intel released 150+ CPU SKUs (unique identifiers for processors)
in 2022 alone~\citep{intel_sku},
each with a unique base frequency, core count, thread count, etc.
The number of possible ways to assemble a system grows exponentially
as we multiply the available options for the various components: CPU, GPU, RAM, display etc.
Computer architects have the challenging job of perfecting a device assembly
to match a target performance goal.
This demands deep domain knowledge, heuristics to estimate performance of designs,
and then extensive testing of prototypes to verify that the desired performance is achieved.
Yet, the manufacturers' claims regarding device performance
often does not match the consumers' experiences~\citep{forbes_benchmarks}.

While there has been prior effort~\citep{li2009accurate,wang2019predicting,tousi2022comparative} towards predicting performance,
they have primarily targeted popular CPU microbenchmark suites,
such as Geekbench~\citep{geekbench} and SPEC~\citep{henning2006spec}.
\citet{li2009accurate} propose an approach based on
Multiple Additive Regression Trees (MARTs)~\citep{friedman2001greedy}
for predicting processor performance on SPEC benchmarks.
\citet{wang2019predicting} use deep neural networks (DNNs) to predict
the performance of Intel CPUs on both Geekbench and SPEC suites.
In a recent work, \citet{cengiz2023predicting} also leverage DNNs to predict
SPEC scores with very high accuracy.
A recent survey by \citet{tousi2022comparative} compares a number of different approaches
for predicting SPEC scores and note that tree-based models provide the best results.
Beyond CPU performance prediction,
\citet{morlans2022power} recently proposed ML models for predicting power consumption.
These approaches demonstrate the power of ML for performance prediction,
but they are too specific to particular subsystems, such as CPU or power,
and do not generalize to predicting overall user experience (UX).

While these microbenchmarks are excellent at stress testing a system,
revealing the ``peak'' system performance,
they do not satisfactorily model the ``average'' UX,
as observed by \citet{forbes_benchmarks}.
Designing workloads that mimic end-user use cases is challenging,
and the large diversity in operating systems and software libraries
makes a fair comparison of UX even more difficult.
In this study, we restrict our attention to ChromeOS~\citep{chromeos},
which primarily runs web applications (webapps),
such as Google Docs, Google Meet, YouTube etc.
Every webapp runs within the Chrome browser, which simplifies the tooling
required for a fair aggregation and comparison of UX indicators.

\begin{table}[t]%
    \centering%
    \renewcommand{\arraystretch}{1.225}%
    \begin{tabular}{ p{11.25em} c p{11.25em} c p{11.25em} }
        \toprule
          \ccol{\textbf{Latency}} & &
          \ccol{\textbf{Responsiveness}} & &
          \ccol{\textbf{Smoothness}} \\
        \midrule
          \ccol{Startup Time} & &
          \ccol{Janky Intervals} & &
          \ccol{Dropped Frames} \\
          {\setstretch{0.95}\small
            Time (ms) since an application invocation to a window launch\par} & &
          {\setstretch{0.95}\small
            Number of 100ms intervals in which a user event was in queue\par} & &
          {\setstretch{0.95}\small
            Fraction (\%) of frames dropped during scrolling or update\par} \\[-1.1em]
        \cmidrule{1-1} \cmidrule{3-3} \cmidrule{5-5}
          \ccol{Tab Switch Time} & &
          \ccol{Key Press Delay} & &
          \ccol{Window Animation} \\
          {\setstretch{0.95}\small
            Time (ms) since a tab switch event to the first rendered frame\par} & &
          {\setstretch{0.95}\small
            Time (ms) taken by application to start a key press event\par} & &
          {\setstretch{0.95}\small
            Relative (\%) FPS (compared to 60 FPS) during window hiding\par} \\[-1.1em]
        \cmidrule{1-1} \cmidrule{3-3} \cmidrule{5-5}
          \ccol{Largest Contentful Paint} & &
          \ccol{Mouse Press Delay} & &
          \ccol{Tab Switch Animation} \\
          {\setstretch{0.95}\small
            Time (ms) taken to paint the largest image or text block\par} & &
          {\setstretch{0.95}\small
            Time (ms) taken by application to start a mouse press event\par} & &
          {\setstretch{0.95}\small
            Relative (\%) FPS (compared to 60 FPS) during tab switching\par} \\[-1.1em]
        \bottomrule
    \end{tabular}\\[0.75em]
    \caption{A subset of UX metrics from the ``User Metrics Analysis'' (UMA) framework in Chrome.}%
    \label{tab:metrics}%
    \vspace*{-1.25em}%
\end{table}

ChromeOS powers the Chromebook~\citep{chromebook} laptops
that are widely used in the education sector~\citep{ahlfeld2017device, williamson2019new, carter2018preparing}.
We define automated tests that replicate common end-user workloads from telemetry data,
including document editing and web browsing,
YouTube playback, and video calling in Google Meet.
We observe that system-level health metrics such as CPU/RAM usage, system load etc.
do not always indicate perceivable UX degradation.
Instead, we identify a subset of metrics from the Chrome browser~\citep{uma},
that capture noticeable performance degradation.
For instance, users notice when an application takes too long to start,
or when an application is slow in responding to keyboard or mouse actions.
Our metrics are part of the Web Vitals~\citep{web_vitals} initiative
that outlines a set of key performance indicators (KPIs) for web applications.
We list these metrics in \cref{tab:metrics},
and discuss more in \cref{sec:data}.

We run our test automation to track 9 UX metrics
on \nChromebooks\ Chromebooks from \nManufacturers\ manufacturers,
and gather \nDataPoints\ UX data points.
We train a set of gradient boosted regression trees (GBRTs)~\citep{friedman2001greedy},
one per UX metric.
We observe an average $R^2$ score~\citep{wright1921correlation} of \meanRScore\
indicating that the trained GBRTs are a ``good fit'' on our dataset,
i.e., they explain the variations in the dataset well.
We also achieve an average MAAPE of \meanMAAPE,
indicating a low percentage error rate in predictions.
We discuss these models and our results in more detail in \cref{sec:eval}.

In summary, the key contributions presented in this paper are the following:

\begin{enumerate}
    \item We define automated tests for ChromeOS webapps
          that mimic end-user workloads,
          and identify a subset Chrome browser metrics
          that strongly correlate with perceivable UX degradation.
    
    \item On \nChromebooks\ Chromebooks,
          we evaluate these UX metrics across our tests,
          and curate a hardware specifications $\mapsto$ UX metrics dataset
          with \nDataPoints\ data points.
    
    \item We train gradient boosted regression trees
          that predict these UX metrics accurately.
\end{enumerate}

\section{Data Collection}%
\label{sec:data}

In this section, we detail our dataset --- our collection of Chromebook devices,
the target workloads and automated tests, the UX metrics we track during these automated tests across all our devices.

\paragraph{Devices \& Specifications.}

We setup a test bed of \nChromebooks\ Chromebooks.
To ensure good diversity,
we collected devices from \nManufacturers\ well-known manufacturers,
containing system on chips (SoCs) from \nSOCVendors\ well-known vendors.
While all our device SoCs have a 64-bit architecture,
we have a mix of both ARM64 and x64 SoCs.
In \cref{sec:app-specs}, we present a distribution of the hardware specs within our test bed.

We exclude the following major components from our current study:
\begin{andlist}
    \item battery specs:
          since we only evaluate UX when devices are on AC power
    \item storage (HDD/SSD) specs:
          since ChromeOS workloads are primarily CPU-intensive~\citep{boroumand2018google}
    \item GPU specs:
          all of our Chromebooks have integrated GPUs and exact specs on core count, FLOPS etc.
          are not readily available in most cases.
\end{andlist}

\paragraph{Workloads \& Tests.}

Our target workloads are inspired by end-user telemetry data.
They primarily focus on web browsing, document editing in Google Docs,
audio\,/\,video calling in Google Meet, and video playback in YouTube.
We defined automated tests that mimic these use cases.

We take several steps to reduce variability in measurements:
\begin{andlist}
    \item our devices run the same ChromeOS version
    \item tests are run only when the device is on AC power
    \item we run each test is run multiple times per device
    \item if a test does not run to completion, we discard all collected metrics.
\end{andlist}

\paragraph{User-Experience Metrics.}

We track 9 metrics, listed in \cref{tab:metrics},
from the ``User Metrics Analysis'' framework~\citep{uma} in the Chrome browser,
that directly quantify UX degradation.
Many of our metrics are part of the Web Vitals~\citep{web_vitals} initiative
that outlines a set of key performance indicators for webapps:

\begin{itemize}
    \item \sfemph{Largest Contentful Paint} (LCP) time~\citep{chrome_lcp} is part of the core set,
    \item A variant of janky intervals~\cite{janky_intervals}, called \sfemph{Total Blocking Time} (TBT) is part of the lab set, and
    \item A key press and mouse press delay are part of the core set as \sfemph{Interaction to Next Paint} (INP).
\end{itemize}

Beyond these page-level metrics, we also track application-level metrics such as startup time
and window\,/\,tab animation smoothness to capture a holistic view of user experience.

As a sanity check before building predictive models,
we also performed a correlation analysis between our Chromebook specifications
and the UX metric values,
to check if user experience generally improved with better hardware.
We discuss the correlation matrix in \cref{sec:app-corr}.

\section{Methodology}%
\label{sec:methodology}

In this section, we overview
\begin{andlist}
    \item our data pipeline --- our data cleaning and feature engineering processes
    \item our machine-learning (regression) model.
\end{andlist}
Formally, for each metric $m$ we train a function $f_m$
that given a vector $\mathbf{x}$ of hardware specifications
predicts an estimated value $\hat{y}_m$ for the metric $m$
(with some true value $y_m$)
based on the model parameters $\hat{\beta}$ learned during training.

\vspace{-1em}%
$$
  \hat{y}_m = f_m(\mathbf{x}, \hat{\beta})
  \qquad\qquad\textrm{where}\ 
  \mathbf{x} = \langle
    x_\textrm{\tiny cpu\_freq},
    \ldots,
    x_\textrm{\tiny ram\_capacity},
    \ldots,
    x_\textrm{\tiny display\_res}
  \rangle
$$

\paragraph{Data Cleaning.}

We discard metrics from tests that terminate before completion.
We also remove extreme values,
such as 100\% dropped frames and 0\% smoothness.
To reduce the impact of outliers, we only use the median of the multiple per-test iterations.
Mean seemed highly susceptible to outliers.

\paragraph{Feature Engineering.}

We list our features \cref{sec:app-features};
all  are numeric, except the following:

\begin{itemize}
    \item We use a one-hot encoding~\citep{zheng2018feature} for the CPU vendor name:
          a categorical feature with 4 choices.
    \item We use a single pixel count feature
          as opposed to using vertical and horizontal components of display resolution,
          as performance typically depends on the number of pixels not their position.
\end{itemize}

\paragraph{ML Model.}

Tree-based models have been shown to be more effective~\citep{chen2016xgboost,shwartz2022tabular}
than neural models,
especially at regression on tabular data, i.e., data with a fixed set of features.
Tabular data often poses challenges for neural networks:
lack of locality, data sparsity, mixed feature types etc.
Further, given their determinism and interpretability,
we use \sfemph{Gradient Boosted Regression Trees} (GBRTs)~\citep{friedman2001greedy}.
We clip the outputs of our GBRTs from below at zero,
since all our metrics assume non-negative values.

\section{Evaluation}%
\label{sec:eval}

We use a 80\% / 10\% / 10\% pseudo-random split of our devices
for training / validation / testing,
ensuring good diversity of specs in all sets.
We detail our training process and then our results below.

\paragraph{Training.}

For training our GBRT models,
we use the \texttt{GradientBoostingRegressor} interface
in the \texttt{scikit-learn}~\citep{scikit-learn} Python library.
We minimize the default loss function on the training data ---
the \sfemph{Mean Squared Error} (MSE)
using \citet{friedman2001greedy}'s least-squares improvement criterion.

We use cross-validated grid search~\citep{muller2016introduction}
to optimize 5 hyperparameters:
\begin{andlist}
    \item learning rate
    \item the number of estimators
    \item subsampling for estimators
    \item depth of estimators
    \item number of features to split on.
\end{andlist}
The optimal hyperparameter values are listed in the appendix, \cref{sec:app-hyper}.

\begin{figure}[t]%
    \hspace{-0.9em}%
    \includegraphics[width=1.04\textwidth]{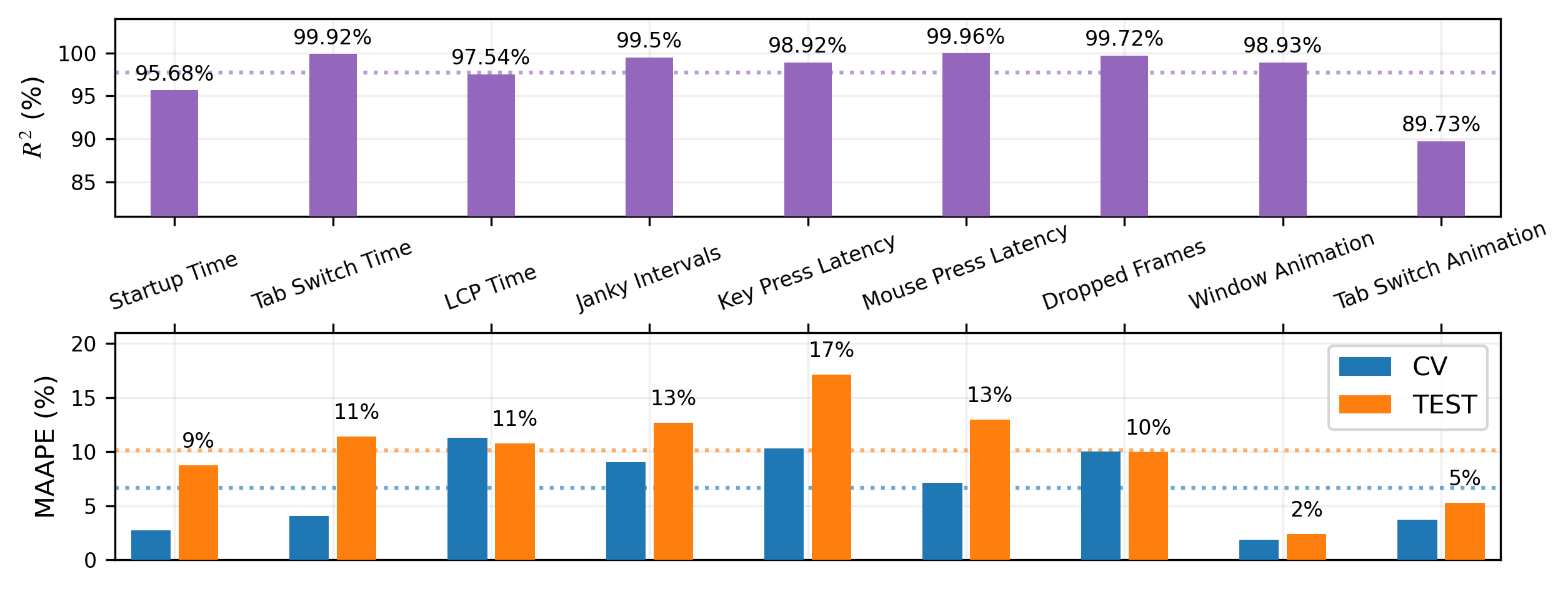}%
    \vspace*{-0.875em}%
    \caption{The $R^2$ fits and MAAPE errors of our predictors.
             Dotted lines indicate the mean values.}%
    \label{fig:accuracy}%
    \vspace{-0.375em}%
\end{figure}

\paragraph{Assessment.}

In \Cref{fig:accuracy}, we present an evaluation of our GBRTs,
one per metric from \cref{tab:metrics}.
We measure
\begin{andlist}
    \item the goodness of fit on learning dataset
    \item the prediction error rates on test dataset.
\end{andlist}

We compute the $R^2$ score~\citep{wright1921correlation},
to measure how well our regression models fit our datasets.
We observe a mean $R^2$ score of \meanRScore,
indicating that our models fit our datasets very well.
However, $R^2$ does not indicate prediction accuracy or model generalization
on \textit{out-of-sample} points~\citep{hawinkel2023out}.
For prediction error rate,
we compute the \sfemph{Mean Arctangent Absolute Percentage Error}
(MAAPE)~\citep{kim2016new},
which provides a stable relative error even when the true values are zero.
For instance, a zero value for the ``Janky Intervals'' metric indicates that
there was no queuing in browser event processing during a test,
which is actually observed on several devices.




In \cref{sec:feat-imp},
we also discuss the permutation importance of features~\citep{breiman2001random} for each GBRT.

\section{Limitations}%
\label{sec:limitations}

While we believe that our results are promising,
we are aware of the following key limitations:
 
\begin{itemize}
    \item While ChromeOS supports Linux and Android applications
          inside virtual machines or within containers,
          we restrict the workloads in our experiments to only native (web) applications.

    \item There is an observation cost in measuring UX metrics while running the workloads.
          We believe this overhead is minimal,
          as we directly use the Chrome browser's UMA~\cite{uma} framework.

    \item We collect UX metrics only when the devices are connected to AC power,
          so our current study makes no claims regarding the user experience
          when laptops are on battery power.
\end{itemize}



\begin{ack}
We thank Jesse Barnes and Dossym Nurmukhanov 
for their support in making this experiment possible.
We also thank Willis Kung and Sanghwan Moon
for their insights and valuable discussions,
and David Lo for reviewing our early drafts and providing feedback. 
\end{ack}

\bibliography{references}

\newpage

\setcounter{section}{0}

\renewcommand\thesection{\Alph{section}}
\renewcommand\thesubsection{\thesection.\arabic{subsection}}

\section{Appendix}%
\label{sec:appendix}

\subsection{Distribution of Hardware Specifications}%
\label{sec:app-specs}

All \nChromebooks\ devices in our test bed had 60 Hz display refresh rate.
Below, we present the distribution of other CPU, RAM and display specs across our devices.

\hspace{-0.375em}%
\includegraphics[width=1.0125\textwidth]{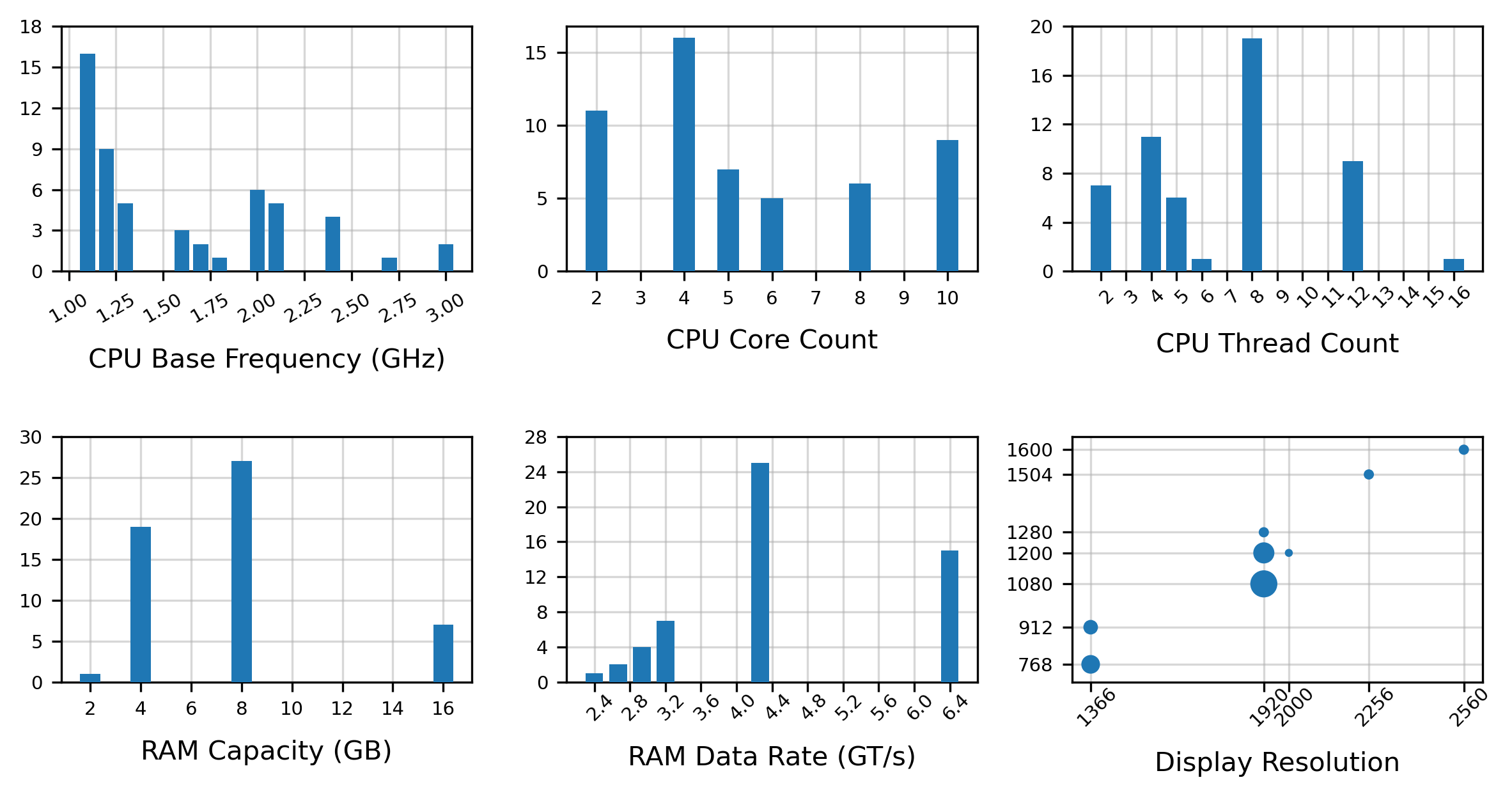}

\subsection{UX Metrics $\sim$ Hardware Specs Correlation}%
\label{sec:app-corr}

As opposed to Pearson's $\rho$ ~\citep{freedman2007statistics},
which captures linear correlation,
we found Kendall's $\tau$ rank correlation~\citep{kendall1938new} to be more informative.
We present the correlation matrix below.

\hspace{-0.35em}%
\includegraphics[width=1.0275\textwidth]{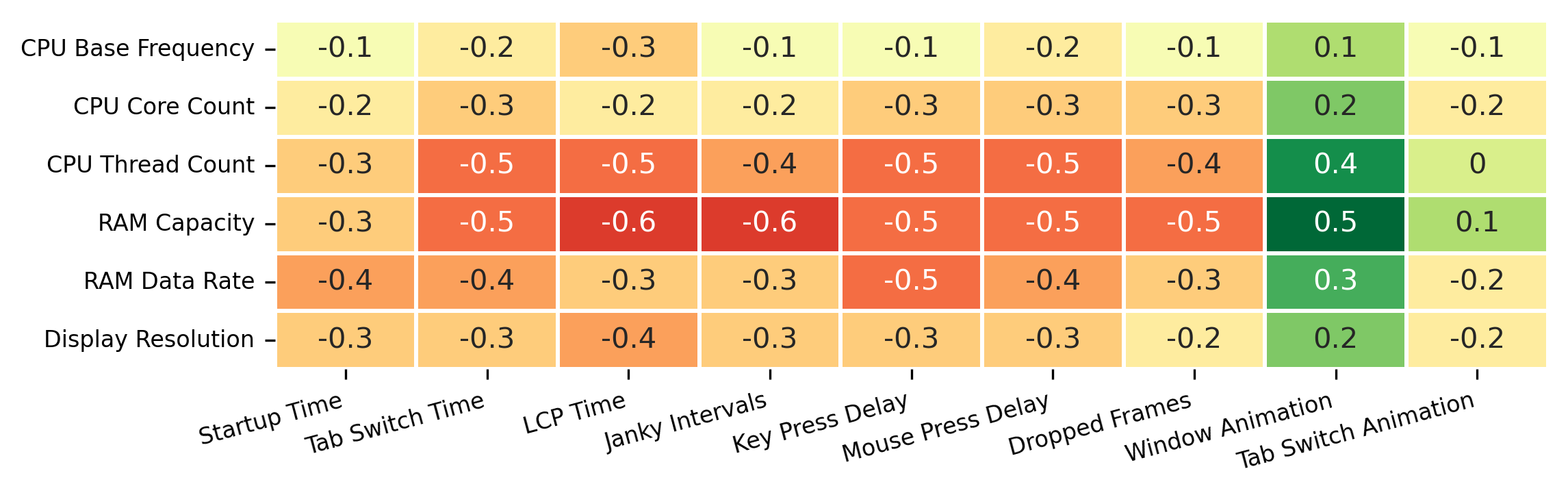}

A positive value indicates that
the UX metric often increases as a hardware spec value improves and vice versa.
Similarly, a negative value indicates that
the UX metric often decreases as a hardware spec value improves and vice versa.
The correlation patterns match our expectations:
\begin{andlist}
    \item latencies, and timings metrics negatively correlate with hardware specs
    \item frame drop rate and jankiness also negatively correlate
    \item animation smoothness metrics positively correlate with hardware specs.
\end{andlist}
We observe that RAM capacity and CPU thread count
have the strongest correlation with most UX metrics.
CPU base frequency seemed to have the weakest correlation across metrics,
but that is expected since the OS dynamically adjusts the CPU's running frequency.

We note one interesting correlation that ran counter to our intuition.
Higher display resolution seems to correlate with \textit{lower} latencies
and with \textit{higher} animation smoothness!
On performing a cross correlation between hardware specifications,
it was revealed that most devices with higher display resolution
simultaneously also had better CPU and RAM specifications,
therefore demonstrating lower latencies and higher animation smoothness.

\subsection{Device Specification Features}%
\label{sec:app-features}

\begin{center}
    \renewcommand{\arraystretch}{1.33}
    \begin{tabular}{ p{10em} p{8em} p{8em} }
        \toprule
          \ccol{\textbf{CPU}} &
          \ccol{\textbf{RAM}} &
          \ccol{\textbf{Display}} \\
        \midrule
          Base Frequency (GHz) &
          Data Rate (GT/s) &
          Pixel Count\,\textsuperscript{*}
          \\[-0.5em]
          {\small (\texttt{Float})} &
          {\small (\texttt{Integer})} &
          {\small (\texttt{Integer})}
          \\
          Core Count &
          Capacity (GB) &
          \st{Refresh Rate (Hz)}\,\textsuperscript{\$}
          \\[-0.5em]
          {\small (\texttt{Integer})} &
          {\small (\texttt{Integer})} &
          {\small (\texttt{Integer})}
          \\
          Thread Count & &
          \\[-0.5em]
          {\small (\texttt{Integer})} & &
          \\
          Vendor Code\,\textsuperscript{\#} &
          &
          \\[-0.5em]
          {\small (\texttt{One-Hot})} & &
          \\
        \bottomrule
    \end{tabular}
\end{center}

\textsuperscript{*} Pixel Count = Horizontal Pixels $\times$ Vertical Pixels,
e.g. 2073600 for 1920 $\times$ 1080 resolution.

\textsuperscript{\$} Refresh rate, although collected,
is dropped from training, since it is 60 Hz for all our devices.

\textsuperscript{\#} Vendor Code is a one-hot encoding~\citep{zheng2018feature}
for 4 distinct CPU vendor names (a categorical field).

\subsection{Model Hyperparameters}%
\label{sec:app-hyper}

In the tables below, an unspecified (-) indicates the following:
\begin{itemize}
    \item for \textsf{Max Features} parameter:
          all features are considered for the best split.
    \item for \textsf{Max Depth} parameter:
          tree nodes are expanded until all leaves are pure.
\end{itemize}

\subsubsection{GBRTs for Latency Metrics}

\begin{center}
    \renewcommand{\arraystretch}{1.05}
    \begin{tabular}{ p{7em} p{5em} p{5em} p{5em} }
        \toprule
          &
          \ccol{\textbf{Startup}} &
          \ccol{\textbf{Tab Switch}} &
          \ccol{\textbf{LCP}} \\[-0.1em]
          &
          \ccol{\textbf{Time}} &
          \ccol{\textbf{Time}} &
          \ccol{\textbf{Time}} \\
        \midrule
          \ccol{\textbf{Estimators}} &
          \ccol{128} &
          \ccol{128} &
          \ccol{96} \\
          \ccol{\textbf{Learning Rate}} &
          \ccol{0.3} &
          \ccol{0.3} &
          \ccol{0.2} \\
          \ccol{\textbf{Subsample}} &
          \ccol{0.7} &
          \ccol{0.7} &
          \ccol{0.5} \\
          \ccol{\textbf{Max Features}} &
          \ccol{3} &
          \ccol{-} &
          \ccol{6} \\
          \ccol{\textbf{Max Depth}} &
          \ccol{2} &
          \ccol{-} &
          \ccol{-} \\
        \bottomrule
    \end{tabular}
\end{center}

\subsubsection{GBRTs for Responsiveness Metrics}

\begin{center}
    \renewcommand{\arraystretch}{1.05}
    \begin{tabular}{ p{7em} p{5em} p{5em} p{5em} }
        \toprule
          &
          \ccol{\textbf{Janky}} &
          \ccol{\textbf{Key Press}} &
          \ccol{\textbf{Mouse Press}} \\[-0.1em]
          &
          \ccol{\textbf{Intervals}} &
          \ccol{\textbf{Delay}} &
          \ccol{\textbf{Delay}} \\
        \midrule
          \ccol{\textbf{Estimators}} &
          \ccol{128} &
          \ccol{96} &
          \ccol{128} \\
          \ccol{\textbf{Learning Rate}} &
          \ccol{0.3} &
          \ccol{0.3} &
          \ccol{0.3} \\
          \ccol{\textbf{Subsample}} &
          \ccol{1.0} &
          \ccol{0.7} &
          \ccol{0.6} \\
          \ccol{\textbf{Max Features}} &
          \ccol{3} &
          \ccol{6} &
          \ccol{3} \\
          \ccol{\textbf{Max Depth}} &
          \ccol{5} &
          \ccol{-} &
          \ccol{5} \\
        \bottomrule
    \end{tabular}
\end{center}

\subsubsection{GBRTs for Smoothness Metrics}

\begin{center}
    \renewcommand{\arraystretch}{1.05}
    \begin{tabular}{ p{7em} p{5em} p{5em} p{5em} }
        \toprule
          &
          \ccol{\textbf{Dropped}} &
          \ccol{\textbf{Window}} &
          \ccol{\textbf{Tab}} \\[-0.1em]
          &
          \ccol{\textbf{Frames}} &
          \ccol{\textbf{Animation}} &
          \ccol{\textbf{Animation}} \\
        \midrule
          \ccol{\textbf{Estimators}} &
          \ccol{96} &
          \ccol{128} &
          \ccol{96} \\
          \ccol{\textbf{Learning Rate}} &
          \ccol{0.3} &
          \ccol{0.1} &
          \ccol{0.3} \\
          \ccol{\textbf{Subsample}} &
          \ccol{1.0} &
          \ccol{1.0} &
          \ccol{0.6} \\
          \ccol{\textbf{Max Features}} &
          \ccol{7} &
          \ccol{-} &
          \ccol{2} \\
          \ccol{\textbf{Max Depth}} &
          \ccol{5} &
          \ccol{5} &
          \ccol{3} \\
        \bottomrule
    \end{tabular}
\end{center}

\subsection{Feature Importance for GBRTs}%
\label{sec:feat-imp}

For a high-level understanding of our models,
we compute the permutation importance of features~\citep{breiman2001random} for each GBRT.
Permutation importance of a particular feature is defined as the decrease in a model score
when only that feature is randomly shuffled.
Below, we show the feature importance matrix
with the feature importance values for each model normalized to $[0,1]$.

\hspace{-0.6em}%
\includegraphics[width=1.025\textwidth]{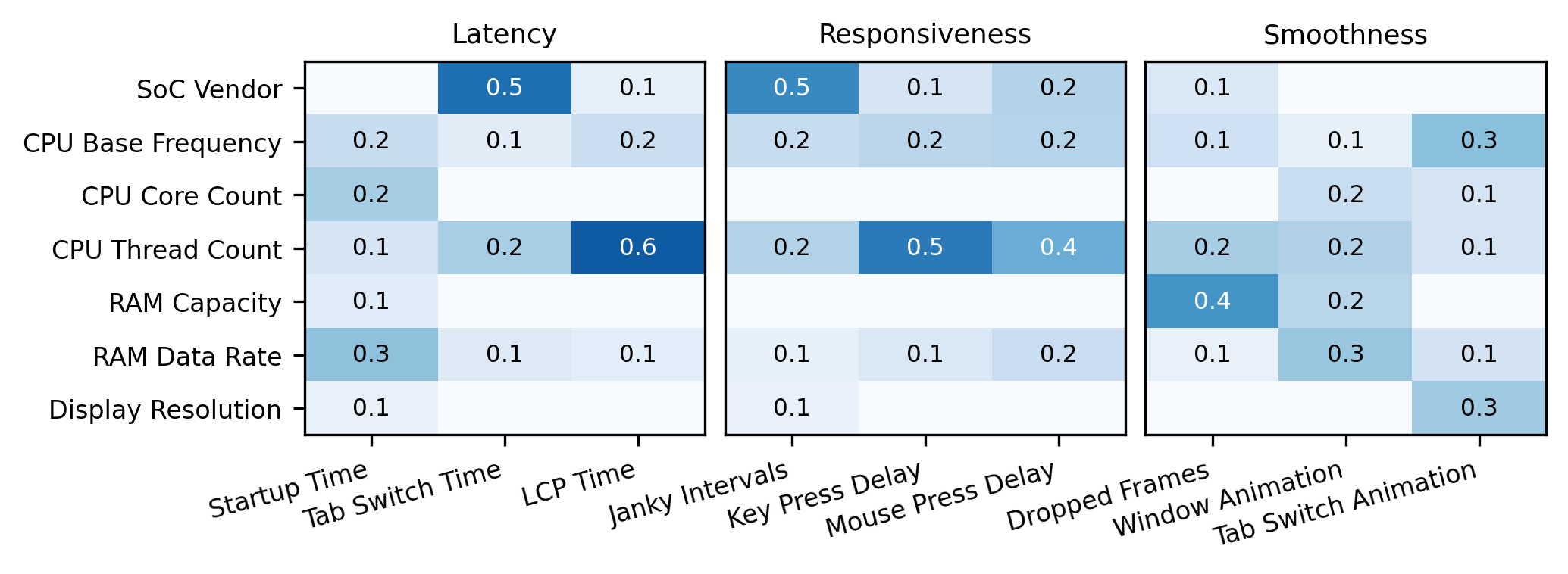}

As in the correlation matrix in \cref{sec:app-corr},
we observe that CPU thread count is one of the most important features across all models.
On the other hand, CPU core count and display resolution
have low importance across all models.
RAM capacity seemed more important in predicting smoothness metrics,
but not so much for latency or responsiveness.
A few models relied heavily on the one-hot encoded SoC vendor id,
perhaps to effectively partition the feature space
and learn more accurate vendor-specific predictors.
It is important to note that unlike correlation analysis,
which reveals true patterns in the data,
feature importance only indicates the subset of specs
that the GBRTs use to accurately partition the feature spaces
and predict metric values with minimal errors.

\end{document}